\documentclass[lettersize, journal]{IEEEtran}  
\usepackage{amsmath,amssymb,amsfonts}
\usepackage{graphicx}
\usepackage{algpseudocode}
\usepackage[linesnumbered,ruled,vlined]{algorithm2e}
\usepackage{pifont}
\usepackage{hyperref}
\usepackage{comment}
\usepackage{multirow}
\usepackage{todonotes}
\usepackage{soul}

\IEEEoverridecommandlockouts

\title{\LARGE \bf PRISM-TopoMap: Online Topological Mapping with Place Recognition and Scan Matching}

\author{Kirill Muravyev$^{1, 2}$, Alexander Melekhin$^{3}$, Dmitry Yudin$^{3,4}$, and Konstantin Yakovlev$^{1,4}$
	\thanks{$^{1}$FRC CSC RAS, Moscow, Russia}%
	\thanks{$^{2}$HSE University, Moscow, Russia}%
	\thanks{$^{3}$MIPT, Dolgoprudny, Russia}%
	\thanks{$^{4}$AIRI, Moscow, Russia}%
	\thanks{\emph{Corresponding author} Kirill Muravyev, \texttt{muraviev@isa.ru} }%
	
}

\begin{document}

	\maketitle
	\thispagestyle{empty}
	\pagestyle{empty}
	
	\begin{abstract}
		
		Mapping is one of the crucial tasks enabling autonomous navigation of a mobile robot. Conventional mapping methods output a dense geometric map representation, e.g. an occupancy grid, which is not trivial to keep consistent for prolonged runs covering large environments. Meanwhile, capturing the topological structure of the workspace enables fast path planning, is typically less prone to odometry error accumulation, and does not consume much memory. Following this idea, this paper introduces PRISM-TopoMap -- a topological mapping method that maintains a graph of locally aligned locations not relying on global metric coordinates. The proposed method involves original learnable multimodal place recognition paired with the scan matching pipeline for localization and loop closure in the graph of locations. The latter is updated online, and the robot is localized in a proper node at each time step. We conduct a broad experimental evaluation of the suggested approach in a range of photo-realistic environments and on a real robot, and compare it to state of the art. The results of the empirical evaluation confirm that PRISM-Topomap consistently outperforms competitors computationally-wise, achieves high mapping quality and performs well on a real robot. The code of PRISM-Topomap is open-sourced and is available at: \url{https://github.com/kirillMouraviev/prism-topomap}.
		
	\end{abstract}
	
	\section{INTRODUCTION}
	
	Building an accurate map of the environment is crucial for mobile robots navigation. Common mapping methods such as RTAB-Map~\cite{labbe2019rtab} or Cartographer~\cite{hess2016real} build maps as dense metric structures like occupancy grids or point clouds. However, such dense metric maps require significant memory for maintenance and optimization, which can potentially lead to memory overflow when the robot navigates large environments~\cite{muravyev2022evaluation}. Coupled with odometry error accumulation, this may cause mapping and loop closure failures with map size growth.
	
	Topological mapping offers an alternative approach to map construction. By capturing just the topological properties of the environment (i.e., connectivity), it significantly reduces memory consumption and computational costs for map maintenance. Additionally, it helps to mitigate mapping failures caused by the odometry error accumulation. Furthermore, the sparsity of topological maps, typically represented as graphs, facilitates faster path planning~\cite{blochliger2018topomap}.
	
\begin{figure}
		\vspace{0.15in}
		\centering
		\includegraphics[width=1\linewidth]{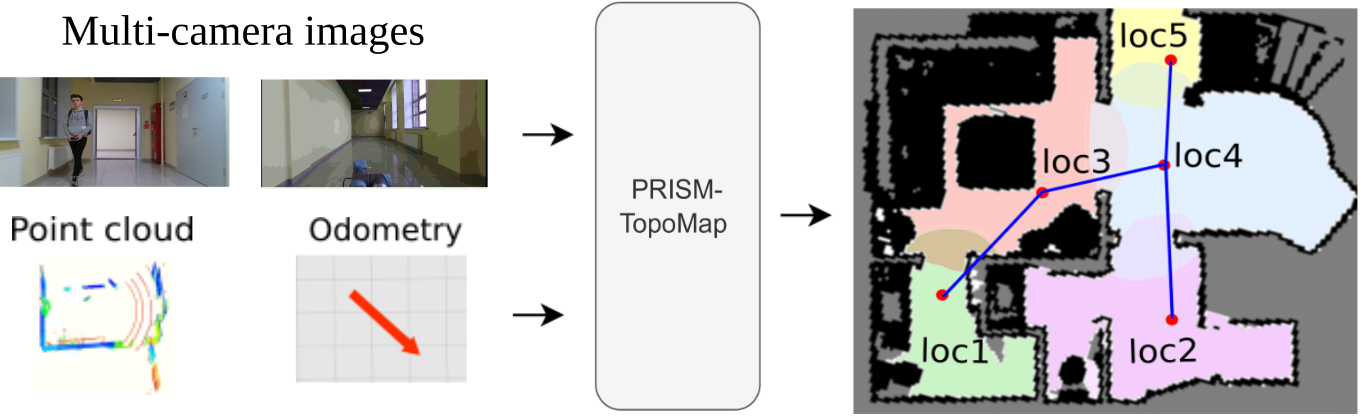}
		\caption{A graph of locations generated from the sensory inputs. Different colors mark different locations, blue lines denote transitions between them.}
		\label{fig:graph_of_locations}
	\end{figure}

	One of the most common topological structures for navigation is a graph of locations (e.g. room, corridor, hall, etc.), where edges denote adjacency or straight-line reachability between the locations -- see Fig.~\ref{fig:graph_of_locations}. Such graphs enable fast and efficient path planning, however, localization in the graph -- i.e., aligning a robot with a location -- is challenging. To address this challenge, learning-based place recognition techniques are typically used nowadays~\cite{yin2022GeneralPlaceRecognition}. However, place recognition errors may lead to creating edges in a graph between non-adjacent locations, which may cause navigation failures. For example, state-of-the-art topological method TSGM~\cite{kim2023topological} links locations at opposite ends of the environment, as shown in~\cite{muravyev2023evaluation}.
	
	To this end we propose a novel approach to building a graph of locations that combines place recognition with 2D feature-based filtering and fine alignment of the identified locations -- PRISM-TopoMap. Our key contributions are as follows:
	\begin{enumerate}
		\item A novel online topological mapping method named PRISM-TopoMap, which builds upon a fine-tuned place recognition model and a point cloud matching technique, and relies solely on local odometry data to localize in the topological map, without using a global metric map or coordinates. These properties allow the proposed method to construct lightweight and consistent graphs of locations with no positioning error accumulation.
		\item A place recognition method for topological mapping, developed through the modification and fine-tuning of the multimodal MSSPlace model~\cite{mssplace}. The improved model named MSSPlace-G effectively integrates multi-camera images and point cloud data, improving the accuracy of place recognition within topological maps.
		\item An original point-cloud matching technique based on 2D features extracted from the cloud projections, which filters place recognition results and estimates relative poses between locations.
	\end{enumerate}
	
	PRISM-Topomap is evaluated and compared with state-of-the-art mapping methods in the photo-realistic Habitat simulator, and is further tested on a real wheeled robot. All experiments involve navigation through large indoor environments. In simulation, PRISM-Topomap constructs consistent topological maps and notably outperforms its competitors across a range of computational performance measures. On a real robot, PRISM-TopoMap also creates a consistent topological map of the whole environment.
	
	\section{RELATED WORK}
	\label{sec:related_work}
	
	\subsection{Topological Mapping}
	
	A large family of methods builds a topological map either based on a pre-built global metric map {\cite{blochliger2018topomap},\cite{chen2022fast}} or together with a global metric map {\cite{hughes2022hydra}}, {\cite{yuan2019incrementally}}. Such hybrid methods provide complete scene representation and enable fast path planning. However, they remain resource-demanding and are susceptible to mapping errors due to odometry error accumulation.
	
	In recent years, a large number of learning-based topological mapping methods that do not rely on a global metric map have emerged {\cite{kim2023topological}}, {\cite{kwon2021visual}}. These methods typically localize within the map relying on neural network-predicted visual embeddings only, without applying filtering to place recognition results. This approach can cause linking distant locations in the environment and result in navigation failures. Other learning-based methods, such as {\cite{wiyatno2022lifelong}}, incorporate edge filtering but require a pre-built topological map at the start.
	
	To better analyze the landscape of the available topological mapping methods, we distinguish several key features:
	
	\begin{itemize}
		\item \textit{Online} -- The method is able to construct a map from scratch in real-time in an online fashion.
		\item \textit{No metric map usage} -- The method does not operate a global metric map or global metric coordinates (so it is not affected by odometry error accumulation).
		\item \textit{Connectivity} -- The method guarantees the connectivity of a built graph.
		\item \textit{Edge filtering} -- The method utilizes a dedicated technique to validate/prune false matching provided by the place recognition module. Mark \ding{55} in this column means that the method relies on matching visual embeddings only.
	\end{itemize}
	
	Table {\ref{tab:properties}} illustrates how state-of-the-art topological mapping methods support these features. The proposed method, PRISM-TopoMap, supports all of them in order to construct consistent graphs in an online mode without accumulating positioning errors.
	
	\subsection{Place Recognition}
	
	Place recognition is crucial for autonomous navigation, enabling robots to recognize previously visited locations. Methods fall into three categories: camera-based, LiDAR-based, and multimodal, combining camera and LiDAR.
	
	Camera-based research highlights include the use of convolutional neural networks (CNNs) for appearance-invariant descriptors~\cite{chen_ConvolutionalNeuralNetworkbased_2014}, the introduction of a trainable VLAD layer in NetVLAD~\cite{arandjelovic2016NetVLADCNNArchitecturea}, and innovations like CosPlace~\cite{berton2022RethinkingVisualGeolocalizationa} which frames learning as a classification challenge. MixVPR~\cite{ali-bey2023MixVPRFeatureMixinga} demonstrates the recent shift towards Transformer architectures in vision tasks.
	Some approaches~\cite{hausler2021PatchNetVLADMultiScaleFusion,wang2022TransVPRTransformerBasedPlace} adopt a two-stage process to refine initial results by focusing on local image regions' descriptors, albeit at a slower computational pace.
	LiDAR-based techniques leverage geometrical data, enhancing robustness. PointNetVLAD~\cite{uy2018PointNetVLADDeepPoint} creates global descriptors from point-wise features. MinkLoc3D~\cite{komorowski2021MinkLoc3DPointCloud} and its enhancement~\cite{komorowski2022ImprovingPointCloudb} use voxelization and sparse convolutions. SVT-Net~\cite{fan2022SVTNetSuperLightWeight} further refines this with an attention mechanism.
	Multimodal methods, like the combination of PointNetVLAD and ResNet50 in~\cite{xie2020LargeScalePlaceRecognition}, leverage both sensor types for richer descriptors, often employing fusion techniques. MinkLoc++~\cite{komorowski2021MinkLocLidarMonoculara} pairs MinkLoc3D with ResNet18 for image and point cloud processing. AdaFusion~\cite{lai2022AdaFusionVisualLiDARFusion} introduces a middle fusion strategy using an attention mechanism for feature integration.
	
	In our paper, we utilize the multimodal and multisensor late-fusion method MSSPlace-G, a modification of MSSPlace~\cite{mssplace}. Additionally, we compare various place recognition models for feature extraction from each modality.
	
	\begin{table}
		\centering
		\caption{Properties of topological mapping methods}
		\setlength\tabcolsep{2pt}
		\scriptsize
		\label{tab:properties}
		\begin{tabular}{c|c|c|c|c}
			\hline
			Method & Online & \begin{tabular}{c}No metric \\ map usage\end{tabular} & Connectivity & \begin{tabular}{c}Edge\\ filtering\end{tabular}\\
			\hline
			Topomap \cite{blochliger2018topomap} & \ding{55} & \ding{55} & \checkmark & \checkmark\\[3pt]
			Taichislam \cite{chen2022fast} & \ding{55} & \ding{55} & \ding{55} & \checkmark\\[3pt]
			Hydra \cite{hughes2022hydra} & \checkmark & \ding{55} & \ding{55} & \checkmark\\[3pt]
			IncrementalTopo \cite{yuan2019incrementally} & \checkmark & \ding{55} & \ding{55} & \checkmark\\[3pt]
			Lifelong \cite{wiyatno2022lifelong} & \ding{55} & \checkmark & \checkmark & \checkmark\\[3pt]
			VGM \cite{kwon2021visual} & \checkmark & \checkmark & \checkmark & \ding{55}\\[3pt]
			TSGM \cite{kim2023topological} & \checkmark & \checkmark & \checkmark & \ding{55}\\[3pt]
			\hline
			(Ours) PRISM-TopoMap & \checkmark & \checkmark & \checkmark & \checkmark\\
			\hline
		\end{tabular}
	\end{table}

	\section{THE TOPOLOGICAL MAPPING PROBLEM}
	\label{sec:problem_statement}
	
	Consider a robot equipped with a perception sensor (e.g. an RGB-D camera or LiDAR) and an odometry sensor. The robot moves through an indoor environment along a predefined trajectory and is tasked with constructing and continuously updating a graph of the visited locations, which can be used for further mission planning and navigation (the latter problems are not addressed in this paper).
	
	We consider 2D environments in this work for the sake of simplicity (however, the suggested methods can function in 3D environments as well). The robot's workspace, $W \subset \mathbb{R}^2$, is composed of the free space and the obstacles: $W = W_{free} \cup W_{obs}$.  
	A location $loc \subset W$ is a subset of the workspace that may represent a room, a corridor segment, a hall, or similar regions (see Fig.~\ref{fig:graph_of_locations} for an example). Each location is associated with its observation point $loc_{obs}$ and its feature map $F(loc) = FMap(loc_{obs})$, which is constructed from the observation captured in this point, e.g. a descriptor extracted from the point cloud using a neural network. Two locations $loc$ and $loc'$ are considered adjacent if they overlap in traversable space: $loc \cap loc' \cap W_{free} \neq \emptyset$. Two adjacent locations define an edge, $e=(loc,\ loc')$. An edge is also associated with its feature map $F(e)$, which is generally used to facilitate navigation between the locations. For instance, a feature map could represent the direction from the observation point of the first location to that of the second location.
	
	\begin{figure*}[ht]
		\vspace*{0.15in}
		\centering
		\includegraphics[width=0.75\textwidth]{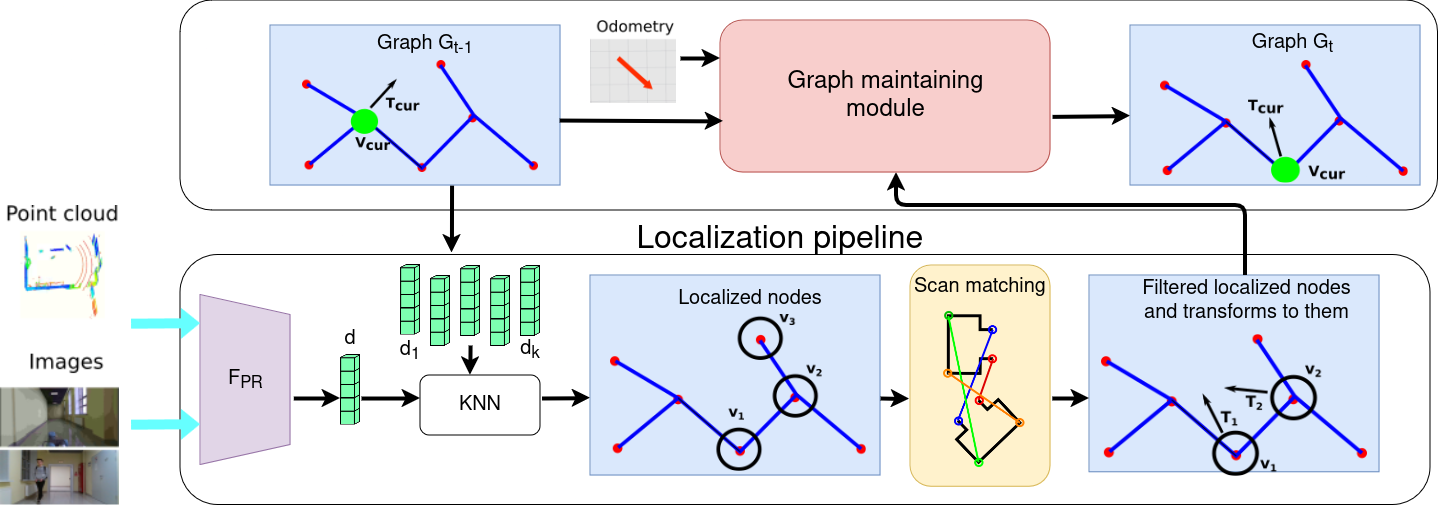}
		\caption{A scheme of the proposed PRISM-TopoMap method which takes multi-camera images and point cloud as input. It includes $F_{PR}$ place encoder, scan matching module, and graph maintaining module. The output is the graph of locations $G_t$ at the moment $t$.}
		\label{fig:method_scheme}
	\end{figure*}
	
	For metric and topometric mapping methods that utilize global metric coordinates (e.g. Hydra {\cite{hughes2022hydra}} or ORB-SLAM3 {\cite{campos2021orb}}), the accuracy of the constructed map/trajectory is typically evaluated. However, for methods that construct a topological map as a graph of locations without global metric coordinates, measuring accuracy becomes problematic and different ways to assess how well the graph of locations represents the environment should be used. For example, the quality of the map in this case may be assessed via the metrics that are directly related to navigation efficiency, such as Success Weighted by Path Length (SPL). Also, the efficiency of navigation within the constructed graph depends on its connectivity and the scene coverage. Based on these considerations, we evaluate the quality of the constructed map, represented as a graph of locations $G=(V,E)$, using the following metrics:
	
	\begin{enumerate}
		\item Connectivity: the number of connected components in the graph.
		\item Coverage of the main connected component: 
		$$Coverage = \frac{Area(\cup_{loc \in V_{main}} loc)}{Area(W)},$$
		where $(V_{main}, E_{main})$ is the main connectivity component (i.e. the component with the largest coverage) of the graph $G$, and $W$ is the robot's workspace.
		\item Effectiveness of path planning with the graph, which is calculated as the SPL value averaged across $N$ sampled pairs $(s_i, g_i)$ of the robot's trajectory points:
		$$SPL = \sum\limits_{i=1}^N I_{cons}(Path_G(s_i, g_i)) \frac{|Path_M(s_i, g_i)|}{|Path_G(s_i, g_i)|},$$
		where $|Path|$ is the length of $Path$; $Path_G(s_i, g_i)$ is the shortest path between points $s_i$ and $g_i$ estimated using the graph $G$; $Path_M(s_i, g_i)$ is the true shortest path between these points; $I_{cons}(Path_G)$ is 1 if $Path_G$ exists and does not contain inconsistent edges, and 0 otherwise. The edge $(u, v)$ is considered inconsistent if it links non-adjacent locations, i.e. $u \cap v \cap W_{free} = \emptyset$.
		
		To estimate the shortest path $Path_G$ between $s_i$ and $g_i$ using the graph $G$, we search across all the locations containing $s_i$, and all the locations containing $g_i$. If the graph has a location which contains both $s_i$ and $g_i$, then $Path_G$ is estimated as the ground truth shortest path between $s_i$ and $g_i$. Otherwise, we choose the path with the minimal length across all $u, v \hookrightarrow g_i \in u, s_i \in v$:
		$$Path_G = \arg\min\limits_{g_i \in u, s_i \in v} |Path(s_i, u, M)| + $$
		$$+ |Path(u, v, G)| + |Path(v, g_i, M)|$$
		
	\end{enumerate}
	
	\section{METHOD}
	\label{sec:method}
	
	A schematic overview of the suggested topological mapping method, PRISM-TopoMap, is provided in Fig.~\ref{fig:method_scheme}. It constructs and updates a graph of locations in the environment in an online mode. To localize within the built graph, it identifies appropriate locations using place recognition, filters the place recognition results, and estimates the robot's position within the identified locations using a 2D feature-based scan matching technique. The method tracks the robot's current state in the graph, which includes the location $v_{cur}^t$ where the robot is currently located, and relative transformation $T_{cur}^t$ between $v_{cur}^t$'s observation point and the robot's position. The state $v_{cur}^t$ and/or $T_{cur}^t$ is updated based on odometry input and localization results. PRISM-TopoMap is comprised of the two main modules: graph maintaining and localization, both of which are described in detail below.
	
	\subsection{Graph Maintaining}
	
	The graph maintaining module builds and expands a graph of locations using observations from the robot's perception sensors and results of the localization module. A location is attributed with the sensory data captured from its observation point (we use 2D projections of LiDAR point clouds and front-view and back-view RGB images). Additionally, each location is assigned a descriptor for place recognition. An edge between locations is attributed with the relative pose between observation points of the locations it connects.
	
	\begin{figure}[ht]
		\centering
		\includegraphics[width=0.38\textwidth]{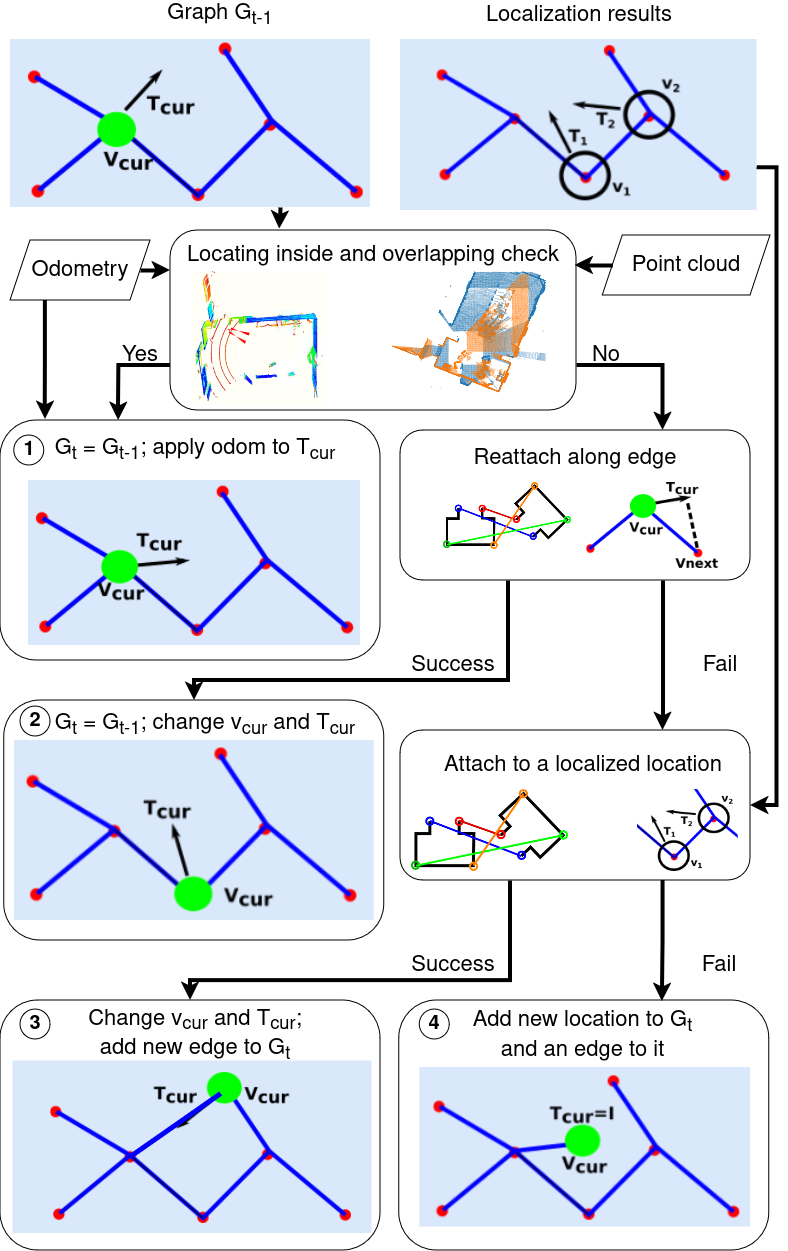}
		\caption{A scheme of the graph maintaining module: checking that the robot is inside $v_{cur}^{t-1}$, changing $v_{cur}$ according to edges and localization results, and addition of new location.}
		\label{fig:graph_maintaining_scheme}
	\end{figure}
	
	At each step $t$, the module outputs the graph of locations $G_t$ covering the area visited by the robot so far, as well as the robot's current state in the graph ($v_{cur}^t, T_{cur}^t$).
	The scheme of the graph maintaining module is depicted in Fig.~\ref{fig:graph_maintaining_scheme}. The input to the module is the graph $G_{t-1} = (V_{t-1}, E_{t-1})$ with state $v_{cur}^{t-1}$ and $T_{cur}^{t-1}$, localization results $V_{loc}$, odometry measurement $o_t$, and point cloud $C_t$ captured from the robot's position. The output of the module is the updated graph $G_t = (V_t, E_t)$ and updated robot state: $v_{cur}^t$ and $T_{cur}^t$. The graph update process is divided into the following stages:
	
	\begin{enumerate}
		\item First, we check whether the robot is still inside $v_{cur}^{t-1}$ and its current scan overlaps with $v_{cur}^{t-1}$ with sufficient percent. If the check passes, we apply the odometry measurement to $T_{cur}$: $v_{cur}^t = v_{cur}^{t-1}; T_{cur}^t = T_{cur}^{t-1} \cdot o_t.$
		
		\item If the robot is outside $v_{cur}^{t-1}$, or scans overlapping percent is low, we first try to change $v_{cur}$'s value to one of its neighbors in the graph (i.e., move along an edge). For this purpose, we match scans of $v_{cur}^{t-1}$ and its neighbors using Harris corner feature detector~\cite{harris1988combined} and the relative transformations assigned to the edges as an initial guess for alignment. If there exist neighbor locations whose scans are matched to $v_{cur}^{t-1}$'s one, we choose the nearest location $v_{next}$ of them and change $v_{cur}$'s value to it:
		$v_{cur}^t = v_{next}; T_{cur}^t = T \cdot T_{cur}^{t-1},$ where $T$ is the transformation from the robot to the observation point of the location $v_{next}$ found by scan matching.
		
		\item Otherwise, we try to change $v_{cur}$'s value to one of the localized locations. For this purpose, we apply the transforms from localization results, and perform an overlap check for each of the localized locations. If there exists a location $v_{loc}$ passing the check, it becomes the current location and is linked to $v_{cur}^{t-1}$:
		$$v_{cur}^t = v_{loc}; T_{cur}^t = T_{loc}; E_t = E_{t-1} \cup \{(v_{cur}^{t-1}, v_{loc})\}.$$ 
		
		\item If there is no proper location to change $v_{cur}$, a new location $v_{new}$ observed from the current robot's position is added to the graph. After addition, we change $v_{cur}$'s value to it and link it to $v_{cur}^{t-1}$ and all the nodes from the localization module output: 
		$v_{cur}^t = v_{new};\ T_{cur}^t = I;$
		$$E_{t} = E_{t-1} \cup \{(v_{new}, v_{cur}^{t-1})\} \cup \{(v_{new}, v_{loc})_{v_{loc} \in V_{loc}}\}.$$
	\end{enumerate}
	
	In all cases of adding a new node, the module links it to the previous location, ensuring the connectivity of the constructed graph. Additionally, switching to an existing location in case (3) automatically closes a loop by adding a new edge. PRISM-TopoMap does not require metric consistency for loops, eliminating the need for resource-intensive global graph optimization.
	
	\subsection{Localization in the Graph via Place Recognition}
	
	MSSPlace method \cite{mssplace} utilizes a late fusion technique to derive a comprehensive global descriptor from the input data, which comprises front-view and back-view images and a point cloud.
	
	\begin{figure}[ht]
		\centering
		\includegraphics[width=0.48\textwidth]{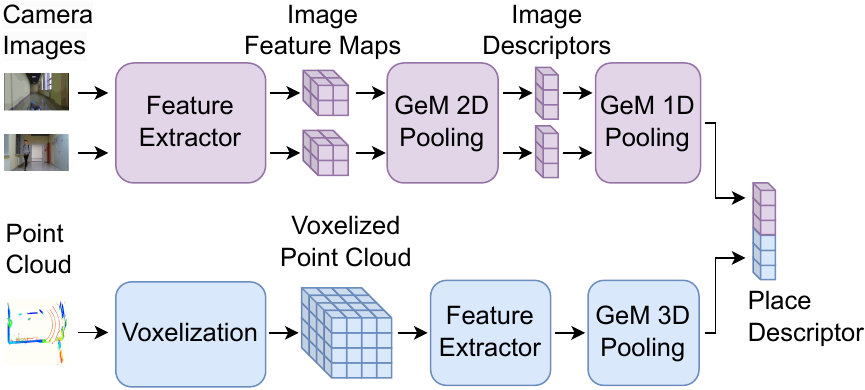}
		\caption{A scheme of the modified MSSPlace place recognition method, referred to as MSSPlace-G.}
		\label{fig:pr_method_scheme}
	\end{figure}
	
	A scheme of the method is shown in Fig. \ref{fig:pr_method_scheme}. MSSPlace-G uses LiDAR point clouds and multiple camera images as input. The point cloud data is initially voxelized and then processed by a sparse 3D convolutional network and the GeM pooling layer to derive the global descriptor for the point cloud, $d_{cloud}$. The images are processed separately by a 2D CNN and GeM pooling. In our modification, we apply a GeM 1D pooling layer to combine the image descriptors into a single global descriptor, denoted as $d_{img}$, rather than summing two image descriptors as in the original MSSPlace. This approach improves performance in generating more effective global descriptors for the images.
	The two descriptors are then concatenated to form a unified descriptor for the current place: $d = Concat(d_{img}, d_{cloud})$.
	
	We then identify the top-5 most similar place descriptors within the nodes of the location graph based on Euclidean distance. These five candidates are utilized in the subsequent stages of our pipeline.
	
	We train the multimodal model in an end-to-end manner, unlike the original MSSPlace, where each modality branch is trained separately. We use the commonly adopted \emph{triplet margin loss}:
	$\mathcal{L}_* = [d_{ap} - d_{an} + m]_+,$
	where $*$ denotes the modality for which the loss is calculated;  $[x]_+$ is the $max(0, x)$ operation; $d_{ap}$ and $d_{an}$ are the Euclidean distances between anchor-positive and anchor-negative pairs of descriptors in the input triplet; and $m$ is the margin hyperparameter. 
	
	To address the dominating modality problem, we compute the loss for image ($\mathcal{L}_{img}$) and point cloud ($\mathcal{L}_{pc}$) modalities separately and calculate the weighted sum:
	\begin{equation}
	\label{eq:loss}
	\mathcal{L} = \alpha\mathcal{L}_{img} + \beta\mathcal{L}_{pc}.
	\end{equation}

	\subsection{Localization Refinement and Filtering via Scan Matching}
	
	Learning-based place recognition methods sometimes output false positive matches due to visual similarity of different parts of the environment. Therefore, relying solely on a place recognition network can cause our topological mapping method to link far, non-adjacent locations. To prevent this, we filter place recognition results matching the scan from the robot with the scans of the found locations, thereby finding relative transforms between these scans.
	
	For fast and robust alignment of scans, we match their 2D projections using features extracted from their geometry. If the matching score for a node is insufficient, we remove this node from the list of localized nodes. Our scan matching algorithm operates on 2D scans, eliminating the need to store point clouds for each location in the graph. This significantly reduces RAM consumption for storing the topological map. While a raw point cloud consumes 1 to 15 MB, depending from sensor resolution, a 2D scan of size 360x360 consumes only about 130 kB.
	
	For point cloud matching, we first remove the floor and ceiling cells from both point clouds. Next, we project both point clouds into 2D and convert the projections to black-and-white image format. After that, we extract features from the resulting images using classical feature detectors like SIFT~\cite{lowe1999object} or ORB~\cite{rublee2011orb}, and match the extracted features using the FLANN method~\cite{muja2009flann}. The resulting transformation is estimated using the least squares method after several iterations of outlier removal. The procedures for outlier removal and transformation estimation are described in Algorithm~\ref{alg:outlier_removal}.
	\begin{algorithm}
		\small
		\caption{Estimation of the transform aligning point clouds and outlier removal}\label{alg:outlier_removal}
		\KwIn{Coordinates of pairwise matched features $M=\{(p_i \in \mathbb{R}^2, q_i \in \mathbb{R}^2)\}_{i=1}^N$; maximum number of iterations $max\_iter$; alignment threshold $\delta$; minimal number of matches $K$}
		\KwOut{Transform $T$ between point clouds}
		\SetKwInOut{Input}{Output}
		\For {$iter=1...max\_iter$} {
			\If {$|M| < K$} {
				return NULL\\
			}
			$T = LeastSquareTransform(M)$\\
			\For {$i=1...N$} {
				\If {$||T p_i - q_i|| > \delta$} {
					$M.remove((p_i, q_i))$\\
				}
			}
		}
		return $LeastSquareTransform(M)$\\
	\end{algorithm}
	
	\section{EXPERIMENTS}
	\label{sec:experiments}
	
	To evaluate PRISM-TopoMap, we tested it on five large simulated scenes and compared it to a range of state-of-the-art metric and topological methods. Furthermore, we conducted experiments using data from a real robot without retraining or fine-tuning any models.
	
	\subsection{Place Recognition Model Training}
	
	The dataset used for training place recognition models and evaluation of our topological mapping method consisted of 180 scenes from HM3D~\cite{ramakrishnan2021habitat}, 73 scenes from Gibson~\cite{xia2018gibson}, and 10 large area scenes from Matterport3D~\cite{chang2017matterport3d}.
	We selected 36, 14 and 5 scenes from each dataset for the validation subset, the rest were used for training.
	Frames were systematically sampled across the scenes from locations positioned on a uniform grid, each spaced 1 meter apart. At each location, four frames were captured, corresponding to rotations of 0, 90, 180, and 270 degrees around the yaw axis.
	
	We followed a standard training approach~\cite{komorowski2021MinkLoc3DPointCloud, komorowski2021MinkLocLidarMonoculara}, utilizing \emph{triplet margin loss} and \emph{batch hard negative mining} to identify challenging triplets within batches, as well as \emph{dynamic batch sizing} to prevent training collapse. In the loss function~\ref{eq:loss}, we set the hyperparameters to $\alpha = 0.5$ and $\beta = 0.5$. Frames within 3 meters were considered a \emph{positive pair}, while those beyond 10 meters were considered \emph{negative pairs}. The learning rate was set to 0.0001 for image-based methods and 0.001 for cloud-based methods. Training was conducted with the Adam optimizer over 60 epochs, with the learning rate reduced by a factor of 0.1 at the 30th and 50th epochs.
	
	\begin{table}[t]
		\caption{The metrics of Place Recognition on the test subset}
		\label{tab:pr_habitat_test}
		\centering
		\setlength\tabcolsep{2pt}
		\begin{tabular}{l|cccc}
			\hline
			Method & Modalities & AR@1\textuparrow & AR@5\textuparrow & Time, ms\textdownarrow \\
			\hline
			GeM \cite{radenovic2018fineGeM} & RGB & 90.09 & 98.50 & 4 \\
			NetVLAD \cite{arandjelovic2016NetVLADCNNArchitecturea} & RGB & 84.29 & 96.60 & 11 \\
			MixVPR \cite{ali-bey2023MixVPRFeatureMixinga} & RGB & 89.77 & 98.58 & 4 \\
			CosPlace \cite{berton2022RethinkingVisualGeolocalizationa} & RGB & 88.67 & 98.08 & 5 \\
			\hline
			MinkLoc3Dv2 \cite{komorowski2022ImprovingPointCloudb} & Point Cloud & 93.16 & 97.79 & 21 \\
			SVT-Net \cite{fan2022SVTNetSuperLightWeight} & Point Cloud & 91.78 & 97.52 & 18 \\
			\hline
			{MinkLoc++} \cite{komorowski2021MinkLocLidarMonoculara} & {RGB + Point Cloud} & 93.96 & 98.18 & 19 \\
			{MSSPlace} \cite{mssplace} & {RGB + Point Cloud} & 95.21 & 98.69 & 27 \\
			{(Ours) MSSPlace-G} & {RGB + Point Cloud} & \textbf{95.26} & \textbf{99.14} & 27 \\
			\hline
		\end{tabular}
	\end{table}
	
	For testing, we evaluated each scene separately by using frames as queries against a database of other frames, excluding frames from identical locations. We calculated Recall@1 and Recall@5 to measure the accuracy of matches within 5 meters for the top-1 and top-5 nearest neighbors. These metrics were averaged across all scenes to compute the Average Recall (AR) metric, as shown in Table~\ref{tab:pr_habitat_test}. For all RGB-only models, we used a ResNet18 backbone. The best metrics were achieved by the MSSPlace-G multimodal and multisensor method. Multimodal methods outperformed unimodal ones (based on RGB-image or Point Cloud) due to the ability of different modalities to complement each other. The MSSPlace and MSSPlace-G methods performed better than MinkLoc++ thanks to the use of multiple camera sensors.
	
	The inference time, reported in Table~\ref{tab:pr_habitat_test}, was measured on a Tesla V100 GPU. MSSPlace-G is slower due to its more complex architecture compared to unimodal methods, as it processes more input data than MinkLoc++, which uses only a single camera. However, the performance remains sufficiently fast for real-time applications.
	
	\subsection{Scan Matching Evaluation}
	
	\begin{table}
		\setlength\tabcolsep{2pt}
		\caption{Results of point cloud matching}
		\centering
		\begin{tabular}{l|cccc}
			\hline
			Method & Precision\textuparrow & Recall@0.5\textuparrow & Recall@0.25 \textuparrow & Runtime, ms\textdownarrow\\
			\hline
			RANSAC + ICP & 0.84 & 0.75 & 0.53 & 360\\
			6x ICP & 0.78 & 0.09 & 0.05 & 580\\
			Geotransformer & 0.22 & 0.85 & 0.67 & 280\\
			\hline
			Feature2D-SIFT & \textbf{1.00} & 0.85 & 0.56 & 75\\
			Feature2D-ORB & \textbf{1.00} & \textbf{0.97} & \textbf{0.69} & \textbf{12}\\
			\hline
		\end{tabular}
		\label{tab:scan_matching_results}
	\end{table}
	
	To determine the most suitable point cloud matching approach for our topological mapping, we compared our 2D feature-based method with several alternatives: a common classical approach combining RANSAC \cite{fischler1981random} and ICP \cite{besl1992method}, bare ICP with six random start positions, and a state-of-the-art learning-based method, GeoTransformer \cite{qin2023geotransformer}.
	
	The point cloud pairs for the evaluation were sampled during the topological mapping process from one of the scenes used in the simulation experiments. In each pair, the first scan was captured from the robot's position at a specific moment, while the second scan corresponded to one of the locations identified by the place recognition pipeline. Please note that the robot was often located several meters away from the observation point of the current location. Additionally, the place recognition pipeline occasionally produced false positives. As a result, the average Intersection over Union (IoU) value among the collected pairs of scans was about 0.25. We used a total of 8,025 point cloud pairs for comparison. Of these, 1,377 pairs had an IoU above 0.5, and 2,369 pairs had an IoU above 0.25. For our approach, we employed SIFT and ORB detectors.
	
\begin{figure}[ht]
		\centering
		\includegraphics[width=0.42\textwidth]{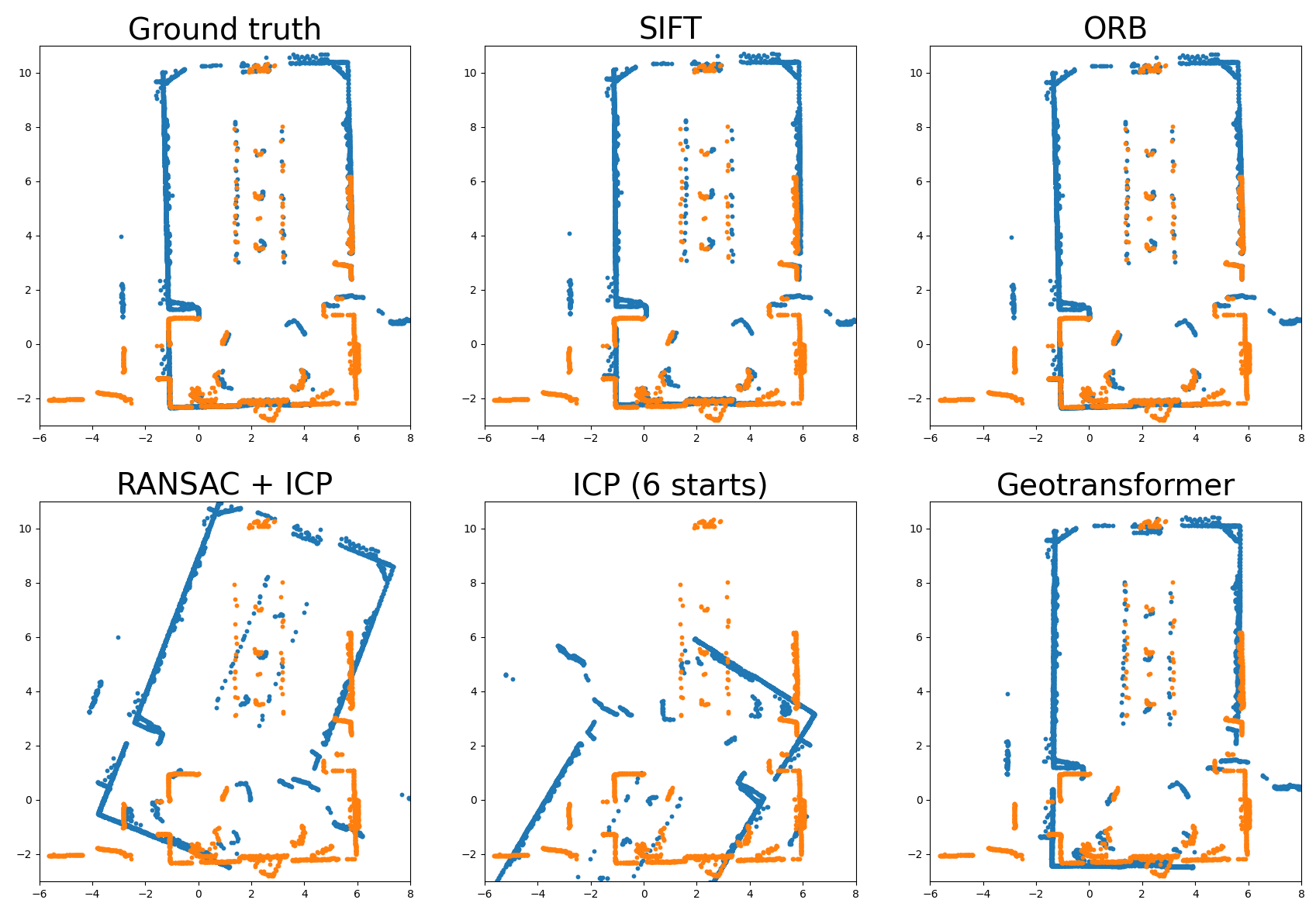}
		\caption{Results of scan matching for all the evaluated methods, compared with ground truth matching. The IoU between scans is 0.65.}
		\label{fig:scan_matching_results}
	\end{figure}

	The results are shown in Table~\ref{tab:scan_matching_results}, and an example of scan matching for all the evaluated methods is illustrated in Fig.~\ref{fig:scan_matching_results}. We estimated precision as the ratio of the number of correctly matched pairs to the total number of pairs matched by a method. For recall estimation, we divided the number of correctly matched pairs by the number of all the pairs with the corresponding overlapping percentage. We considered a pair matched correctly if the error between estimated and ground truth transformation was less than 0.5 m. Runtime was measured on a PC with 6-core CPU and 6 GB RTX 2060 GPU.
	
	As shown in the table, ICP-based methods demonstrated poor recall at the pairs with low overlap, because they search for an optimal transform using the entire point cloud, which may differ significantly from the other point cloud in the pair. Geotransformer reached poor precision because it always considers point clouds as matched. The proposed point cloud matching method using the ORB detector achieved the best inference time, precision, and recall. Therefore, we selected it for both simulation and real robot experiments.
	
	\begin{table*}[ht]
		\caption{Topological mapping metric values on simulation experiments}
		\label{tab:results}
		\centering
		\begin{tabular}{l|l|ccc|ccc|ccc}
			\hline
			& & \multicolumn{3}{c}{Zero noise} \vline & \multicolumn{3}{c}{Medium noise} \vline & \multicolumn{3}{c}{Large noise}\\
			\hline
			& Method & $N_{comp}$\textdownarrow & Coverage\textuparrow & SPL\textuparrow & $N_{comp}$\textdownarrow & Coverage\textuparrow & SPL\textuparrow & $N_{comp}$\textdownarrow & Coverage\textuparrow & SPL\textuparrow\\
			\hline
			Metric & RTAB-Map & \textbf{1.0} & 0.79 & \textbf{0.86} & \textbf{1.0} & 0.81 & \textbf{0.89} & 1.6 & 0.68 & 0.61\\
			& ORB-SLAM3 & 2.2 & \textbf{0.95} & 0.74 & 2.2 & \textbf{0.95} & 0.74 & 2.2 & \textbf{0.95} & 0.74\\
			& GLIM & \textbf{1.0} & 0.80 & 0.72 & \textbf{1.0} & 0.80 & 0.72 & \textbf{1.0} & 0.80 & 0.72\\
			\hline
			Topo & Hydra & 10.0 & 0.77 & 0.38 & 9.6 & 0.80 & 0.34 & 11.8 & 0.83 & 0.41\\
			& IncrementalTopo & 7.6 & 0.89 & 0.66 & 11.4 & 0.74 & 0.38 & 17.2 & 0.43 & 0.20\\
			& \textbf{[Ours]} PRISM-TopoMap & \textbf{1.0} & 0.90 & 0.85 & \textbf{1.0} & 0.92 & 0.87 & \textbf{1.0} & 0.93 & \textbf{0.92}\\
			\hline
		\end{tabular}
	\end{table*}
	
	\subsection{Topological Mapping In Simulation Experiments}
	
	\begin{figure*}[ht]
		\centering
		\vspace*{0.15in}
		\includegraphics[width=1.0\textwidth]{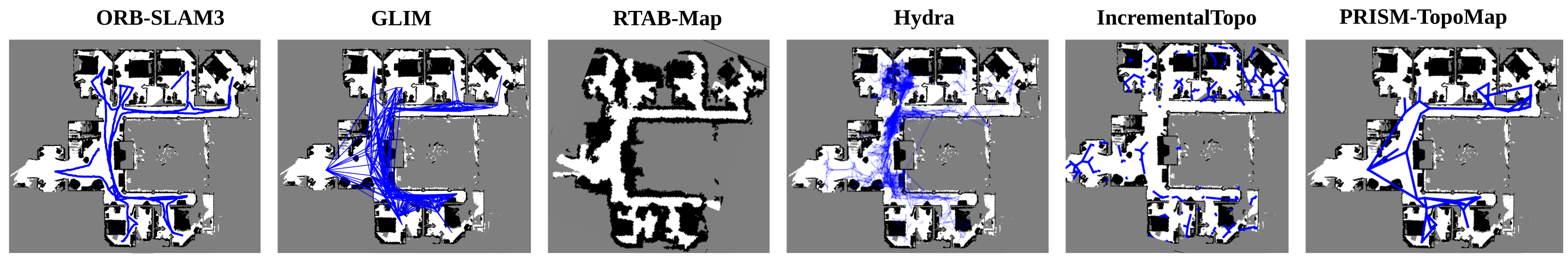}
		\caption{Graphs built on simulated data, aligned with ground truth 2D grid maps, and a metric map built by RTAB-Map method. ORB-SLAM built disconnected graph, as well as Hydra and IncrementalTopo methods. GLIM built dense submap graph with high scene coverage. RTAB-Map built a grid which covers not the whole environment. PRISM-TopoMap method built sparse and connected graph with high scene coverage.}
		\label{fig:built_graphs}
	\end{figure*}
	
	We used five large scenes from the Matterport3D dataset~\cite{chang2017matterport3d} for evaluation. In each scene, a virtual robot moved along a predefined trajectory, exploring the entire scene. These scenes and trajectories are available \href{https://github.com/KirillMouraviev/PRISM-TopoMap/tree/main/data/mp3d_toposlam_validation_scenes}{in the project repo}. The trajectory lengths varied from 100 to 300 meters, and the scene areas ranged from 100 to 700 m$^2$. At each step, the robot received a panoramic RGB-D image from the four cameras with a 90-degree field of view, along with a point cloud generated from this image and odometry data from the simulator. The output of a run was a graph constructed after the agent completed the entire trajectory.
	
	To model real-world conditions, Gaussian noise was added to the odometry every second, separately for the orientation angle and the robot's speed. We used two levels of noise: medium (linear std. 0.003, angular std. 0.0075) and large (linear std. 0.0075, angular std. 0.025), as well as zero noise (i.e., ground truth odometry from the simulator). The medium and large odometry noise reflect an error of state-of-the-art LiDAR-based odometry methods, which is typically between 0.5\% and 1.0\%.
	
	For comparison, we used two state-of-the-art topological mapping methods: Hydra~\cite{hughes2022hydra} and IncrementalTopo~\cite{yuan2019incrementally}, as well as three metric mapping methods: the widely used LiDAR-based method RTAB-Map~\cite{labbe2019rtab}, and state-of-the-art visual (ORB-SLAM3~\cite{campos2021orb}) and LiDAR-based (GLIM~\cite{koide2024glim}) SLAM methods. RTAB-Map required odometry input, which was provided by Cartographer~\cite{hess2016real}. The Hydra and IncrementalTopo methods were supplied with point cloud and odometry data, ORB-SLAM3 was fed with RGB-D data only, and GLIM was provided with point cloud data only. RTAB-Map also received the front image from the robot. For Hydra, semantic data was additionally computed using the SegFormer~\cite{xie2021segformer} neural network. Our method used front and back images, a point cloud, and odometry as input data.
	
	Hydra and IncrementalTopo constructed a global metric map (as a 3D mesh and occupancy grid, respectively) first and then built a topological graph on top of it. GLIM and ORB-SLAM3 generated a sparse global point cloud map. Additionally, GLIM produced a graph of submaps, and ORB-SLAM3 generated a graph of keyframes. To evaluate the ORB-SLAM3 and GLIM methods, we calculated metrics based on these graphs. RTAB-Map created a global metric map in the form of an occupancy grid, which we evaluated as the resulting graph. To estimate SPL for RTAB-Map, we compared the shortest paths in the generated occupancy grid against the ground truth shortest paths.
	
	The results of the comparison are shown in Table~\ref{tab:results}. An example of the built maps for all the methods, using large odometry noise, is shown in Fig.~\ref{fig:built_graphs}. The Hydra and IncrementalTopo methods produced disconnected maps. ORB-SLAM3 failed to localize in three scenes, leading to keyframe graph disconnection. GLIM failed in one scene, covering only 25\% of its area. RTAB-Map generated connected maps with zero and medium positioning noise but produced a disconnected graph in one scene with large positioning noise. However, even with precise positional input, RTAB-Map did not cover the entire scene, with an average coverage percentage of 0.8. In contrast, our method built connected graphs with high coverage (above 0.9 on average) for all the scenes. Moreover, the coverage rate and SPL values for our method remained consistent as positioning noise increased, while RTAB-Map and IncrementalTopo methods showed a significant drop in both coverage and SPL under high noise conditions.
	
	To evaluate the computational burden of the methods we measured the amount of RAM required for map maintenance, as well as map updates and loop closure time, and the resulting map size in disk storage. All performance metrics were measured on a PC with a 6-core Intel Core i5-9500F CPU, GeForce RTX 2060 GPU, and 32 GB of RAM. Table~\ref{tab:performance} shows the results for PRISM-Topomap, ORB-SLAM and Hydra. Clearly, PRISM-TopoMap requires significantly less memory for operation and map storage than its competitors (up to orders of magnitude).
	
	Overall, the obtained results confirm that the suggested method notably outperforms competitors computationally-wise, while providing high-quality map representations that are on par quality-wise (or even better in certain aspects) with state-of-the-art SLAM approaches.
	
	\begin{table}
		\caption{Performance metric values on simulation experiments}
		\setlength\tabcolsep{2pt}
		\label{tab:performance}
		\centering
		\begin{tabular}{l|cccc}
			\hline
			Method & \begin{tabular}{c}RAM usage,\\ GB\end{tabular} & \begin{tabular}{c}Map update\\ time, s\end{tabular} & \begin{tabular}{c}Loop closure\\ time, s\end{tabular} & \begin{tabular}{c}Map size,\\ MB\end{tabular}\\
			\hline
			ORB-SLAM3 & 0.60 & 0.13 & 0.25 & 237\\
			Hydra & 2.20 & 0.90 & 2.00 & 195\\
			PRISM-TopoMap & \textbf{0.15} & \textbf{0.11} & \textbf{0.11} & \textbf{0.4}\\
			\hline
		\end{tabular}
	\end{table}
	
	\subsection{Experiments On The Real Robot}
	
	To demonstrate the efficiency of our method in real-world scenarios, we conducted an experiment using a wheeled mobile robot Husky A200, which moved through hallways of a university building. The robot received point clouds from a Velodyne VLP-16 LiDAR and odometry data from the wheel encoders. Since the robot was equipped with only a front-facing camera, we used the MinkLoc3D model for global localization.
	
\begin{figure}[ht]
		\vspace{0.15in}
		\centering
		\includegraphics[width=0.48\textwidth]{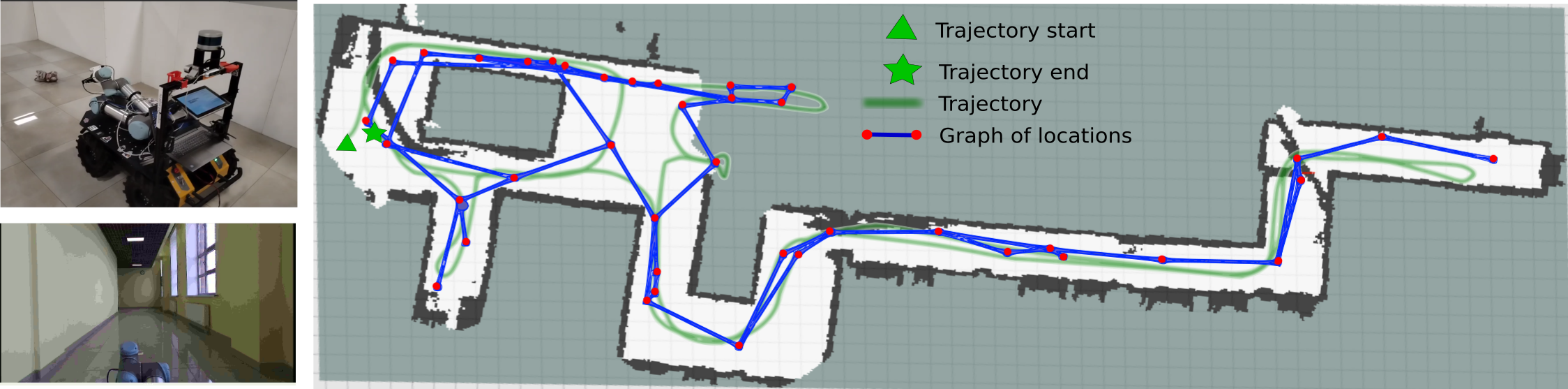}
		\caption{Left: Image of the Husky A200 robot and an image from its front camera. Right: The resulting graph of locations, aligned with global metric occupancy grid built by RTAB-Map (black-and-white), and robot's trajectory (green).}
		\label{fig:results_on_real_data}
	\end{figure}

	The robot traveled a total distance of 212 meters. The result of the topological mapping is shown in Fig.~\ref{fig:results_on_real_data}. Despite the highly noisy wheel odometry and the presence of people walking near the robot, the PRISM-TopoMap method successfully built a graph of locations. A video of the experiment is available at \url{https://youtu.be/aWDK_0L-Vkg}.
	
	\section{CONCLUSION AND FUTURE WORK}
	\label{sec:conclusion}
	
	We have proposed PRISM-TopoMap -- a novel topological mapping method that combines learning-based multimodal place recognition, feature-based scan matching, and rule-based graph maintaining to build connected and consistent graphs of locations from raw perception and odometry sensor data. The experiments conducted on large, photorealistic simulated scenes and on a real wheeled robot demonstrated that the proposed method constructs connected graphs with high scene coverage, even in the presence of severe odometry noise. Furthermore, it significantly outperforms state-of-the-art methods in terms of computational budget. Moving forward, we plan to incorporate semantic information into our method for more accurate localization and to develop navigation techniques using the constructed graphs of locations.
	
	\section*{ACKNOWLEDGMENT}
	
	The reported study was supported by the Ministry of Science and Higher Education of the Russian Federation under Project 075-15-2024-544.
	
	\bibliography{prism-topomap}

\begin{thebibliography}{10}
\providecommand{\url}[1]{#1}
\csname url@rmstyle\endcsname
\providecommand{\newblock}{\relax}
\providecommand{\bibinfo}[2]{#2}
\providecommand\BIBentrySTDinterwordspacing{\spaceskip=0pt\relax}
\providecommand\BIBentryALTinterwordstretchfactor{4}
\providecommand\BIBentryALTinterwordspacing{\spaceskip=\fontdimen2\font plus
\BIBentryALTinterwordstretchfactor\fontdimen3\font minus
  \fontdimen4\font\relax}
\providecommand\BIBforeignlanguage[2]{{%
\expandafter\ifx\csname l@#1\endcsname\relax
\typeout{** WARNING: IEEEtran.bst: No hyphenation pattern has been}%
\typeout{** loaded for the language `#1'. Using the pattern for}%
\typeout{** the default language instead.}%
\else
\language=\csname l@#1\endcsname
\fi
#2}}

\bibitem{labbe2019rtab}
M.~Labb{\'e} and F.~Michaud, ``Rtab-map as an open-source lidar and visual
  simultaneous localization and mapping library for large-scale and long-term
  online operation,'' \emph{Journal of Field Robotics}, vol.~36, no.~2, pp.
  416--446, 2019.

\bibitem{hess2016real}
W.~Hess, D.~Kohler, H.~Rapp, and D.~Andor, ``Real-time loop closure in 2d lidar
  slam,'' in \emph{2016 IEEE international conference on robotics and
  automation (ICRA)}.\hskip 1em plus 0.5em minus 0.4em\relax IEEE, 2016, pp.
  1271--1278.

\bibitem{muravyev2022evaluation}
K.~Muravyev and K.~Yakovlev, ``Evaluation of rgb-d slam in large indoor
  environments,'' in \emph{Interactive Collaborative Robotics: 7th
  International Conference, ICR 2022, Fuzhou, China, December 16-18, 2022,
  Proceedings}.\hskip 1em plus 0.5em minus 0.4em\relax Springer, 2022, pp.
  93--104.

\bibitem{blochliger2018topomap}
F.~Blochliger, M.~Fehr, M.~Dymczyk, T.~Schneider, and R.~Siegwart, ``Topomap:
  Topological mapping and navigation based on visual slam maps,'' in \emph{2018
  IEEE International Conference on Robotics and Automation (ICRA)}.\hskip 1em
  plus 0.5em minus 0.4em\relax IEEE, 2018, pp. 3818--3825.

\bibitem{yin2022GeneralPlaceRecognition}
P.~Yin, S.~Zhao, I.~Cisneros, A.~Abuduweili, G.~Huang, M.~Milford, C.~Liu,
  H.~Choset, and S.~Scherer, ``General {{Place Recognition Survey}}:
  {{Towards}} the {{Real-world Autonomy Age}},'' Sept. 2022.

\bibitem{kim2023topological}
N.~Kim, O.~Kwon, H.~Yoo, Y.~Choi, J.~Park, and S.~Oh, ``Topological semantic
  graph memory for image-goal navigation,'' in \emph{Conference on Robot
  Learning}.\hskip 1em plus 0.5em minus 0.4em\relax PMLR, 2023, pp. 393--402.

\bibitem{muravyev2023evaluation}
K.~Muravyev and K.~Yakovlev, ``Evaluation of topological mapping methods in
  indoor environments,'' \emph{IEEE Access}, vol.~11, pp. 132\,683--132\,698,
  2023.

\bibitem{mssplace}
\BIBentryALTinterwordspacing
A.~Melekhin, D.~Yudin, I.~Petryashin, and V.~Bezuglyj, ``Mssplace: Multi-sensor
  place recognition with visual and text semantics,'' 2024. [Online].
  Available: \url{https://arxiv.org/abs/2407.15663}
\BIBentrySTDinterwordspacing

\bibitem{chen2022fast}
X.~Chen, B.~Zhou, J.~Lin, Y.~Zhang, F.~Zhang, and S.~Shen, ``Fast 3d sparse
  topological skeleton graph generation for mobile robot global planning,'' in
  \emph{2022 IEEE/RSJ International Conference on Intelligent Robots and
  Systems (IROS)}.\hskip 1em plus 0.5em minus 0.4em\relax IEEE, 2022, pp.
  10\,283--10\,289.

\bibitem{hughes2022hydra}
N.~Hughes, Y.~Chang, and L.~Carlone, ``Hydra: a real-time spatial perception
  system for 3d scene graph construction and optimization,'' 2022.

\bibitem{yuan2019incrementally}
Y.~Yuan and S.~Schwertfeger, ``Incrementally building topology graphs via
  distance maps,'' in \emph{2019 IEEE International Conference on Real-time
  Computing and Robotics (RCAR)}.\hskip 1em plus 0.5em minus 0.4em\relax IEEE,
  2019, pp. 468--474.

\bibitem{kwon2021visual}
O.~Kwon, N.~Kim, Y.~Choi, H.~Yoo, J.~Park, and S.~Oh, ``Visual graph memory
  with unsupervised representation for visual navigation,'' in
  \emph{Proceedings of the IEEE/CVF International Conference on Computer
  Vision}, 2021, pp. 15\,890--15\,899.

\bibitem{wiyatno2022lifelong}
R.~R. Wiyatno, A.~Xu, and L.~Paull, ``Lifelong topological visual navigation,''
  \emph{IEEE Robotics and Automation Letters}, vol.~7, no.~4, pp. 9271--9278,
  2022.

\bibitem{chen_ConvolutionalNeuralNetworkbased_2014}
Z.~Chen, O.~Lam, A.~Jacobson, and M.~Milford, ``Convolutional {{Neural
  Network-based Place Recognition}},'' Nov. 2014.

\bibitem{arandjelovic2016NetVLADCNNArchitecturea}
R.~Arandjelovi{\'c}, P.~Gronat, A.~Torii, T.~Pajdla, and J.~Sivic,
  ``{{NetVLAD}}: {{CNN}} architecture for weakly supervised place
  recognition,'' May 2016.

\bibitem{berton2022RethinkingVisualGeolocalizationa}
G.~Berton, C.~Masone, and B.~Caputo, ``Rethinking {{Visual Geo-localization}}
  for {{Large-Scale Applications}},'' \emph{arXiv:2204.02287 [cs]}, Apr. 2022.

\bibitem{ali-bey2023MixVPRFeatureMixinga}
A.~{Ali-bey}, B.~{Chaib-draa}, and P.~Gigu{\`e}re, ``{{MixVPR}}: {{Feature
  Mixing}} for {{Visual Place Recognition}},'' in \emph{Proceedings of the
  {{IEEE}}/{{CVF Winter Conference}} on {{Applications}} of {{Computer
  Vision}}}, 2023, pp. 2998--3007.

\bibitem{hausler2021PatchNetVLADMultiScaleFusion}
S.~Hausler, S.~Garg, M.~Xu, M.~Milford, and T.~Fischer, ``Patch-{{NetVLAD}}:
  {{Multi-Scale Fusion}} of {{Locally-Global Descriptors}} for {{Place
  Recognition}},'' in \emph{Proceedings of the {{IEEE}}/{{CVF Conference}} on
  {{Computer Vision}} and {{Pattern Recognition}}}, 2021, pp. 14\,141--14\,152.

\bibitem{wang2022TransVPRTransformerBasedPlace}
R.~Wang, Y.~Shen, W.~Zuo, S.~Zhou, and N.~Zheng, ``{{TransVPR}}:
  {{Transformer-Based Place Recognition With Multi-Level Attention
  Aggregation}},'' in \emph{CVPR}, 2022, pp. 13\,648--13\,657.

\bibitem{uy2018PointNetVLADDeepPoint}
M.~A. Uy and G.~H. Lee, ``{{PointNetVLAD}}: {{Deep Point Cloud Based
  Retrieval}} for {{Large-Scale Place Recognition}},'' in \emph{CVPR}, 2018,
  pp. 4470--4479.

\bibitem{komorowski2021MinkLoc3DPointCloud}
J.~Komorowski, ``{{MinkLoc3D}}: {{Point Cloud Based Large-Scale Place
  Recognition}},'' in \emph{Proceedings of the {{IEEE}}/{{CVF Winter
  Conference}} on {{Applications}} of {{Computer Vision}}}, 2021, pp.
  1790--1799.

\bibitem{komorowski2022ImprovingPointCloudb}
J.~Komorowski, ``Improving {{Point Cloud Based Place Recognition}} with
  {{Ranking-based Loss}} and {{Large Batch Training}},'' in \emph{2022 26th
  {{International Conference}} on {{Pattern Recognition}} ({{ICPR}})}, Aug.
  2022, pp. 3699--3705.

\bibitem{fan2022SVTNetSuperLightWeight}
Z.~Fan, Z.~Song, H.~Liu, Z.~Lu, J.~He, and X.~Du, ``{{SVT-Net}}: {{Super
  Light-Weight Sparse Voxel Transformer}} for {{Large Scale Place
  Recognition}},'' \emph{AAAI}, vol.~36, no.~1, pp. 551--560, June 2022.

\bibitem{xie2020LargeScalePlaceRecognition}
S.~Xie, C.~Pan, Y.~Peng, K.~Liu, and S.~Ying, ``Large-{{Scale Place Recognition
  Based}} on {{Camera-LiDAR Fused Descriptor}},'' \emph{Sensors}, vol.~20,
  no.~10, p. 2870, Jan. 2020.

\bibitem{komorowski2021MinkLocLidarMonoculara}
J.~Komorowski, M.~Wysocza{\'n}ska, and T.~Trzcinski, ``{{MinkLoc}}++: {{Lidar}}
  and {{Monocular Image Fusion}} for {{Place Recognition}},'' in \emph{2021
  {{International Joint Conference}} on {{Neural Networks}} ({{IJCNN}})}, July
  2021, pp. 1--8.

\bibitem{lai2022AdaFusionVisualLiDARFusion}
H.~Lai, P.~Yin, and S.~Scherer, ``{{AdaFusion}}: {{Visual-LiDAR Fusion With
  Adaptive Weights}} for {{Place Recognition}},'' \emph{IEEE Robotics and
  Automation Letters}, vol.~7, no.~4, pp. 12\,038--12\,045, Oct. 2022.

\bibitem{campos2021orb}
C.~Campos, R.~Elvira, J.~J.~G. Rodr{\'\i}guez, J.~M. Montiel, and J.~D.
  Tard{\'o}s, ``Orb-slam3: An accurate open-source library for visual,
  visual--inertial, and multimap slam,'' \emph{IEEE Transactions on Robotics},
  vol.~37, no.~6, pp. 1874--1890, 2021.

\bibitem{harris1988combined}
C.~Harris, M.~Stephens, \emph{et~al.}, ``A combined corner and edge detector,''
  in \emph{Alvey vision conference}, vol.~15, no.~50.\hskip 1em plus 0.5em
  minus 0.4em\relax Citeseer, 1988, pp. 10--5244.

\bibitem{lowe1999object}
D.~G. Lowe, ``Object recognition from local scale-invariant features,'' in
  \emph{Proceedings of the seventh IEEE international conference on computer
  vision}, vol.~2.\hskip 1em plus 0.5em minus 0.4em\relax Ieee, 1999, pp.
  1150--1157.

\bibitem{rublee2011orb}
E.~Rublee, V.~Rabaud, K.~Konolige, and G.~Bradski, ``Orb: An efficient
  alternative to sift or surf,'' in \emph{2011 International conference on
  computer vision}.\hskip 1em plus 0.5em minus 0.4em\relax Ieee, 2011, pp.
  2564--2571.

\bibitem{muja2009flann}
M.~Muja and D.~Lowe, ``Flann-fast library for approximate nearest neighbors
  user manual,'' \emph{Computer Science Department, University of British
  Columbia, Vancouver, BC, Canada}, vol.~5, p.~6, 2009.

\bibitem{ramakrishnan2021habitat}
S.~K. Ramakrishnan, A.~Gokaslan, E.~Wijmans, O.~Maksymets, A.~Clegg, J.~Turner,
  E.~Undersander, W.~Galuba, A.~Westbury, A.~X. Chang, \emph{et~al.},
  ``Habitat-matterport 3d dataset (hm3d): 1000 large-scale 3d environments for
  embodied ai,'' \emph{arXiv preprint arXiv:2109.08238}, 2021.

\bibitem{xia2018gibson}
F.~Xia, A.~R. Zamir, Z.~He, A.~Sax, J.~Malik, and S.~Savarese, ``Gibson env:
  Real-world perception for embodied agents,'' in \emph{Proceedings of the IEEE
  conference on computer vision and pattern recognition}, 2018, pp. 9068--9079.

\bibitem{chang2017matterport3d}
A.~Chang, A.~Dai, T.~Funkhouser, M.~Halber, M.~Niessner, M.~Savva, S.~Song,
  A.~Zeng, and Y.~Zhang, ``Matterport3d: Learning from rgb-d data in indoor
  environments,'' \emph{arXiv preprint arXiv:1709.06158}, 2017.

\bibitem{radenovic2018fineGeM}
F.~Radenovi{\'c}, G.~Tolias, and O.~Chum, ``Fine-tuning cnn image retrieval
  with no human annotation,'' \emph{IEEE transactions on pattern analysis and
  machine intelligence}, vol.~41, no.~7, pp. 1655--1668, 2018.

\bibitem{fischler1981random}
M.~A. Fischler and R.~C. Bolles, ``Random sample consensus: a paradigm for
  model fitting with applications to image analysis and automated
  cartography,'' \emph{Communications of the ACM}, vol.~24, no.~6, pp.
  381--395, 1981.

\bibitem{besl1992method}
P.~J. Besl and N.~D. McKay, ``Method for registration of 3-d shapes,'' in
  \emph{Sensor fusion IV: control paradigms and data structures}, vol.
  1611.\hskip 1em plus 0.5em minus 0.4em\relax Spie, 1992, pp. 586--606.

\bibitem{qin2023geotransformer}
Z.~Qin, H.~Yu, C.~Wang, Y.~Guo, Y.~Peng, S.~Ilic, D.~Hu, and K.~Xu,
  ``Geotransformer: Fast and robust point cloud registration with geometric
  transformer,'' \emph{IEEE Transactions on Pattern Analysis and Machine
  Intelligence}, 2023.

\bibitem{koide2024glim}
K.~Koide, M.~Yokozuka, S.~Oishi, and A.~Banno, ``Glim: 3d range-inertial
  localization and mapping with gpu-accelerated scan matching factors,''
  \emph{Robotics and Autonomous Systems}, vol. 179, p. 104750, 2024.

\bibitem{xie2021segformer}
E.~Xie, W.~Wang, Z.~Yu, A.~Anandkumar, J.~M. Alvarez, and P.~Luo, ``Segformer:
  Simple and efficient design for semantic segmentation with transformers,''
  \emph{NeurIPS}, vol.~34, pp. 12\,077--12\,090, 2021.

\end{thebibliography}
	\bibliographystyle{myIEEEtran}
	
\end{document}